\documentclass{article}

    \usepackage[preprint]{neurips_2020}

\usepackage{amsmath,amsfonts,bm}

\def\eqref#1{equation~\ref{#1}}

\def\1{\bm{1}}

\DeclareMathAlphabet{\mathsfit}{\encodingdefault}{\sfdefault}{m}{sl}
\SetMathAlphabet{\mathsfit}{bold}{\encodingdefault}{\sfdefault}{bx}{n}

\def\gA{{\mathcal{A}}}

\def\taue{\tau_E}

\DeclareMathOperator*{\argmin}{arg\,min}

\usepackage{xcolor}
\usepackage[utf8]{inputenc} 
\usepackage[T1]{fontenc}    
\usepackage{hyperref}       
\usepackage{natbib}
\usepackage{amsfonts}       
\usepackage{nicefrac}       
\usepackage{microtype}      
\usepackage{graphicx}
\usepackage{caption}
\usepackage{subcaption}
\usepackage[ruled,vlined,noline, noend]{algorithm2e}
\newcommand{\bfx}{\mathbf{x}}
\newcommand{\bfy}{\mathbf{y}}
\title{Towards Improving Learning from Demonstration Algorithms via MCMC Methods}

\author{
  Carl Qi, Edward Sun, Harry Zhang\\
  Course Project for the PGM Course at CMU
}

\begin{document}

\maketitle
\begin{abstract}
Behavioral cloning, or more broadly, learning from demonstrations (LfD) is a priomising direction for robot policy learning in complex scenarios. Albeit being straightforward to implement and data-efficient, behavioral cloning has its own drawbacks, limiting its efficacy in real robot setups. In this work, we take one step towards improving learning from demonstration algorithms by leveraging implicit energy-based policy models. Results suggest that in selected complex robot policy learning scenarios, treating supervised policy learning with an implicit model generally performs better, on average, than commonly used neural network-based explicit models, especially in the cases of approximating potentially discontinuous and multimodal functions.
\end{abstract}
\section{Introduction}
Learning from demonstration (LfD) has been an active topic in robotics research. In the context of robotics and automation, learning from demonstration (LfD) is the paradigm in which robots acquire new skills by learning to imitate an expert. Among a variety of LfD algorithms, behavioral cloning (BC) \cite{pomerleau1998autonomous} remains one of the most straightforward methods to acquire skills in the real world where the robotic agent tries to imitate (i.e. clone) an expert policy via supervised learning. Despite its simple problem formulation and significant inductive bias acquired from supervised learning methods, the method has valid both empirical and theoretical shortcomings such as sample inefficiency and distribution shift in inference time \cite{ross2011reduction, tu2021sample}, in practice it enables some of the most compelling results of real robots generalizing complex
behaviors to new unstructured scenarios \cite{zhang2018deep, florence2019self, zhang2021robots}. The choice of LfD over other robot learning methods is compelling when ideal behavior can be neither easily scripted (as is done in traditional robot programming) nor easily defined as an optimization problem, but can be demonstrated. In this project we tackle the topic from a probabilistic point of view, where we investigate the effect of novel sampling techniques in learning from demonstration. This work is inspired by recent investigation and advancement in the field of LfD using implicit models \cite{avigal2021avplug, du2019, du2020compositional}, where the BC problem is reformulated as an implicit function calculated with an energy-based model via:
\[
\hat{a}=\mathrm{argmin}_{a\in\mathcal{A}}
E_\theta(o,a)\; \mathrm{instead\; of}\; \hat{a}=\pi_\theta(o) .
\]
With the introduction of this implicit energy-based policy, we would need to intelligently sample from this function in order to produce on-policy samples during training and test time. Thus, we investigate how implicit function with different probabilistic sampling methods would improve the usual behavioral cloning/imitation learning performance. We conduct experiments in complex simulated environments that both naive BC and naive RL policy such as DDPG and SAC would fail \cite{haarnoja2018soft, hou2017novel}. Results suggest that this simple change can lead to remarkable improvements in performance across a wide range of contact-rich tasks such as deformable objects manipulation, especially simulated dough manipulation tasks.

The contribution of our work is three-fold:
\begin{enumerate}
    \item A comprehensive evaluation of various sampling methods in LfD algorithms.
    \item Simulated experiments that compare regular policy learning vs. energy-model-based implicit policy learning.
    \item A sufficiently complex simulated platform for deformable objects manipulation that could be used as a test bed for other LfD algorithms.
\end{enumerate}

\section{Related Work}
\textbf{Learning from Demonstrations.} Learning from demonstrations, or often interchangeably, imitation learning, is a prevalent method in the field of robot learning \cite{pomerleau1998autonomous, zhang2016health}. In addition, data-driving methods such as \cite{peng2018deepmimic, ross2011reduction, zhang2020dex} combined with existing BC or RL methods have shown great potential in complex robotic tasks. Generative distribution-matching algorithms such as \cite{ho2016generative, zhang2021robots} require no labeling, but may require millions of on-policy environment interactions, which is expensive to collect in real robot setups. Other methods such as \cite{kumar2020conservative, fu2020d4rl, eisner2022flowbot3d} attempt to leverage both the abundance of pre-collected of offline data and the generalizability and flexibility of online value function-based policy learning in regular RL setup and come up with offline RL algorithms. Offline RL algorithms often show competitive performance that is even superior to expert demonstration performance. Perhaps the success of BC comes from its simplicity: it has the lowest data collection
requirements, can be data-efficient \cite{ross2011reduction, avigal20206}, and is arguably the simplest to implement and easiest to tune due to the lack of hyperparameters.

\textbf{Energy-Based Models and Implicit Policy Learning.} \cite{lecun2006tutorial, song2021train, devgon2020orienting, avigal20206} provide a comprehensive overview of how energy-based models are formulated and trained. Energy-based models are chosen in many scenarios due to the desirable properties of out-of-distribution generalization and long-horizon sequential prediction \cite{du2020compositional, sim2019personalization} and discontinuity modeling \cite{florence2021implicit, elmquist2022art}. When used in robot learning tasks, EBMs with implicit policy learning are shown to outperform traditional explicit policies in a variety of tasks such as contact-rich pushing and placing, deformable objects manipulation, and door opening \cite{florence2021implicit, eisner2022flowbot3d}. Moreover, the flexibility of MCMC sampling and EBMs also facilitates their prevalence in other domains such as computer vision tasks and text generation tasks \cite{gustafsson2020energy, deng2020residual, pan2022tax}. In reinforcement learning, \cite{haarnoja2017reinforcement, zhang2023flowbot++} uses an EBM formulation as
the policy representation. Other recent work \cite{du2020compositional, jin2024multi} uses EBMs in a model-based planning framework, or uses EBMs in imitation learning but with an on-policy algorithm. Another line of recent RL works has emerged to utilize an EBM as part of an overall policy \cite{kostrikov2021offline, nachum2021provable, shen2024diffclip}.

\textbf{Data-driven Deformable Objects Manipulation Methods.} Due to the difficulties with defining a deformable object's state representation, prior works have developed data-driven methods to learn from examples. Prior works mainly use the available data in two ways. One way is to leverage the available data to train a dynamics model and later use it for planning. Some train a particle-based dynamics model~\cite{li2018learning, lin2021learning, yao2023apla}, some train a visual dynamics model~\cite{2020vsf, lim2021planar, lim2022real2sim2real}, and others trains a latent dynamics model~\cite{ma2021learning,Matl2021DeformableEO}. Among the model-based approaches, DPI-Net~\cite{li2018learning, teng2024gmkf} is the most related to modeling elastic/plastic objects like dough.

As deformable object manipulation becomes increasingly popular, the need for training data result in many high-quality simulators \cite{hu2019difftaichi,corl2020softgym, heiden2021disect,huang2021plasticinelab}. We choose PlasticineLab~\cite{huang2021plasticinelab}, which uses Material Point Method~\cite{hu2018moving} to model elastoplastic material. The simulator is built on top of the DiffTaichi system~\cite{hu2019difftaichi}, which supports differentiable physics that allow us to perform gradient-based trajectory optimization.

\begin{figure}[t]
     \centering
         \centering
         \centerline{\includegraphics[width=\textwidth]{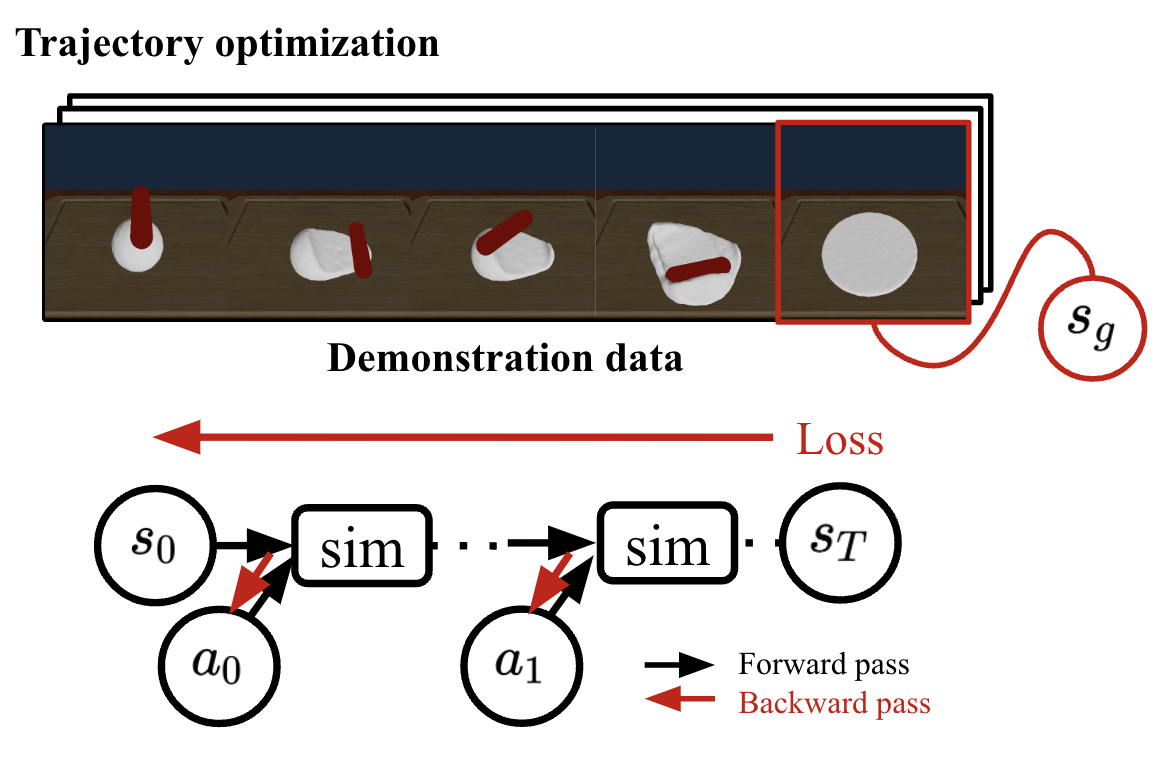}}
        \caption{We perform trajectory optimization to obtain expert demonstrations. We first compute the loss between a target state and the states in our trajectory. We then back-propagate the gradient from the target shape through a differentiable simulator to get the updated actions.}
        \label{fig:traj}
\end{figure}

\section{Background}
\subsection{Markov Decision Process}
A Markov Decision Process (MDP) is defined using the tuple: $\mathcal{M} =
{\mathcal{S}, \mathcal{R}, \mathcal{A}, \mathcal{O},\mathcal{P}, \rho_0, \gamma}$. $\mathcal{S}, \mathcal{A}$, and $\mathcal{O}$ represent the state, action, and observation space. $\mathcal{R} : \mathcal{S}\times\mathcal{A} \rightarrow
\mathcal{R}$ is the reward function. $\mathcal{P} : \mathcal{S}\times\mathcal{A} \rightarrow \mathcal{S}$ is the transition dynamics. $\rho_0$ is the probability distribution
over initial states and $\rho_0 = [0, 1)$ is a discount factor. Let $\pi : \mathcal{S} \rightarrow \mathcal{A}$ be a policy which maps states to actions. In the partially observable case, at each time $t$, the policy maps a partial observation $o_t$ of
the environment to an action $a_t = \pi(o_t)$. Our goal is to learn a policy that maximizes the expected
cumulative rewards $\mathbb{E}_\pi[
\sum_{t=0}^\infty\gamma^tr_t]$, where $r_t$ is the reward at time $t$. The Q-function of the policy
for a state-action pair is $Q(s, a) = \mathcal{R}(s, a) + \gamma \mathbb{E}_{s',\pi}[\sum_{t=0}^\infty\gamma^tr_t|s_0=s']$
where $s'$
represents the next state of taking action a in state s according to the transition dynamics.

\subsection{Imitation Learning}
In imitation learning, we assume access to a set of demonstration trajectories trajectories $\taue \in D$ from an expert.
Every trajectory $\tau = (s_0, a_0, s_1, a_1, \cdots)$ consists of a finite-length sequence of state and action pairs
The goal of imitation learning is to learn a policy $\pi_\theta$ that imitates the expert policy used to generate the demonstration.

\subsection{Behavioral Cloning}
Behavioral cloning (BC) directly leverages supervised learning over state-action pairs in the expert demonstrations. In the deterministic case, the policy is learned by solving a regression problem with states $s_t$ as the features and actions $a_t$ and target labels.
Formally, we minimize the following objective:
\begin{align}
   \ell_{BC}(\theta) := \sum_{(s_t, a_t) \in \taue } \Vert a_t - \pi_\theta(s_t) \Vert_2^2.
\end{align}

Probabilistically, the BC problem can also be framed as optimizing the log-likelihood of the expert trajectories:
\begin{align}
    \ell_{BC}(\theta) := -\log\left(\sum_{i=1}^N P_\theta(\tau_i)\right)
\end{align}

where $P_\theta$ is a (learned) function modeling the probabilistic distribution of the expert demonstration data. Common choice of the $P_\theta$ model is a normal distribution defined as:
\[
P_\theta(s_t, a_t) \propto \mathcal{N}(\mu_\theta(s_t), \sigma_\theta(s_t))
\]

\subsection{Implicit Behavioral Cloning}
\label{bg:ibc}
While the aforementioned methods explicitly solves for the BC objective function to learn a policy that takes as input a state and outputs an action, Implicit Behavioral Cloning \citep{florence2021implicit} formulates BC using implicit models – specifically, the composition of $\argmin$ with a continuous energy function $E_\theta$ to represent the policy $\pi_\theta$:
\begin{align}
    \hat a = \argmin_{a \in \gA} E_\theta (s, a)
\end{align}
This formulates imitation learning as a conditional energy-based modeling (EBM) problem. During training, an energy model over the state and actions $E_\theta(s, a)$ are trained and at inference time (given $s$) performs implicit regression by optimizing for the optimal action $\hat a$ via sampling or gradient descent. Existing literature \citep{florence2021implicit} claims that implicit models outperform explicit model in modeling discontinuity and generalization.
\begin{figure}[t]
  \centering
  \includegraphics[width=0.7\textwidth]{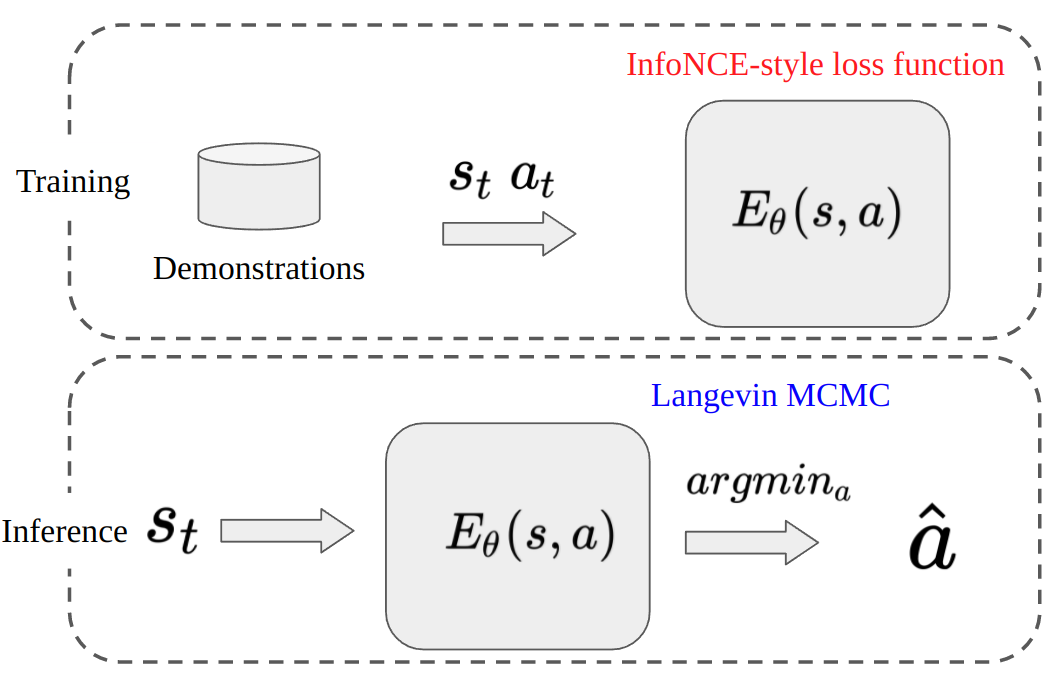}
  \caption{Baseline method: Implicit Behavioral Cloning. The energy-based model over states and actions is trained via an InfoNCE-style loss function. The inference is done by a MCMC sampling-based optimization procedure.}
  \label{fig-main}
\end{figure}
\section{Method}
\subsection{Generate Demonstrations by Trajectory Optimization}
We leverage gradient-based trajectory optimization with a differentiable simulator to generate demonstration data. In Direct Shooting form~\cite{underactuated}, our trajectory optimization with all the constraints captured by the dynamics of the environment is shown below. Given the initial state $s_0$, the objective is:
\begin{flalign}
    \min_{a_1, ..., a_T} &L_{traj} = \sum_{t=0}^T L^{task}_t + \lambda L^{contact}_t \label{eq:loss}\\
    &= \sum_{t=0}^T d(s_t, s_g) + \lambda s_t 
    \label{eq:loss-full}\\
     \text{s.t.} ~& s_{t+1} = \mathcal{T}(s_t, a_t)
\end{flalign}
 where $s_t$, is the current state,  $s_g$ is the goal state, $d$ is the similarity between the two, $L^{contact}$ is a contact loss, and $\lambda$ weighs the different loss terms. With differentiable dynamics, we can compute the gradient of the loss with respect to the actions:
\begin{align}
    \frac{\partial L_{traj}}{\partial a_t} = \sum_{t'=t+1}^T \frac{\partial \left(L^{task}_t + \lambda L^{contact}_t\right)}{\partial s_{t'}} \frac{\partial s_{t'}}{\partial a_t}
    \label{eq:gradient_updates}
\end{align} 
Here, $\frac{\partial s_{t'}}{\partial a_t}$ can be computed using the back-propagation technique with a differentiable dynamics model. Therefore, a straightforward approach used by many prior works~\cite{huang2021plasticinelab, li2018learning, qi2022dough} is to use gradient-descent to directly update the actions, as shown in Figure~\ref{fig:traj}.

\subsection{Energy-based Policy Learning }
As described in Sec.~\ref{bg:ibc}, we define a energy-based policy as the composition of $\arg\min$ with a continuous energy function $E_{\theta}$.

Given a dataset of expert demonstration $\{o_i, a_i\}$, where $s_i$ and $a_i$ are the observation and action at time step i, we use Noise Contrastive Estimation (NCE) to train the energy model. We assume the actions are n-dimensional and bounded by $a_{\text{min}}, a_{\text{max}} \in \mathbb{R}^n$. More specifically, we follow recent literature in energy model training and use InfoNCE \cite{oord2018representation} loss. 

InfoNCE loss (Eq.~\ref{eq:infonce}) can be seen as using cross-entropy loss to identify the positive action amongst a set of noise samples. To construct a training pair, we sample a tuple of demonstration $(o_i, a_i)$ from the dataset, where $a_i$ is the expert action. Then we sample $N_\text{neg}$ negative actions ${\color{red}\{\tilde{\mathbf{y}}^j_i\}_{j=1}^{N_{\text{neg.}}}  }$ drawn from some proposal distribution. The specific type of proposal distribution will described in the later sections. $e^{-E_{\theta}( \mathbf{o}_i, {\color{black} \mathbf{a}_i})} +  {\color{red} \sum_{j=1}^{N_{\text{neg}}}} e^{-E_{\theta}(\mathbf{o}_i, {\color{red} \tilde{\mathbf{a}}^j_i} )}$ can be seen as an approximation of the partition function $Z(\mathbf{o}_i, \theta)$, and larger $N_\text{neg}$ always leads to more stable training.
\begin{equation}
    \mathcal{L}_{\text{InfoNCE}} = \sum_{i=1}^N -\log \big( \tilde{p}_{\theta}( {\color{black} \mathbf{a}_i} | \ \mathbf{o_i}, \ {\color{red}\{\tilde{\mathbf{a}}^j_i\}_{j=1}^{N_{\text{neg.}}}  } ) \big)
\label{eq:infonce}
\end{equation}

\begin{equation}
    \tilde{p}_{\theta}( {\color{black} \mathbf{a}_i} | \ \mathbf{o_i}, \ {\color{red}\{\tilde{\mathbf{a}}^j_i\}_{j=1}^{N_{\text{neg.}}}  } ) =  \frac{e^{-E_{\theta}(\mathbf{o}_i, {\color{black} \mathbf{a}_i}  )}} {e^{-E_{\theta}( \mathbf{o}_i, {\color{black} \mathbf{a}_i})} +  {\color{red} \sum_{j=1}^{N_{\text{neg}}}} e^{-E_{\theta}(\mathbf{o}_i, {\color{red} \tilde{\mathbf{a}}^j_i} )} }
\end{equation}

In the following sections, we will describe two methods for inference and the corresponding proposal distribution for negative samples during training. 

\subsubsection{Gradient-free Optimization}
\textbf{Training}. We assume the actions are bounded and sample the negative samples from uniform distribution
${ \tilde{\mathbf{a}}} \sim \mathcal{U}(\mathbf{a}_{\text{min}}, \mathbf{a}_{\text{max}})$
, where $\mathbf{a}_{\text{min}}, \mathbf{a}_{\text{max}}\in \mathbb{R}^m$. To compute  $\mathcal{L}_{\text{InfoNCE}}$, we use a batch size (N) of 100 with 256 negative examples per sample.

\textbf{Inference}. During test-time, given a trained energy model $E_{\theta}(\bfx, \bfy)$, we use a simple gradient-free optimization method to get the action with lowest energy. The algorithm below is very similar to Cross-entropy Method (CEM) except that we use Expectation-Maximization (EM) algorithm to fit a Gaussian Mixture Model (GMM) to capture the multimodality of optimal action.

\begin{algorithm}[H]
 Initialize: $\{\tilde{\bfy_i}\}_{i=1}^{N_{\text{samples}}} \sim \mathcal{U}(\mathbf{a}_{\text{min}}, \mathbf{a}_{\text{max}})$, $\sigma = \sigma_{\text{init}}$  \;
  \For{iter in 1 to $N_{\text{iters}}$}{
  
    $\{E_i\}_{i=1}^{N_{\text{samples}}} = \{E_{\theta}(\bfx, \tilde{\bfy}_i)\}_{i}^{N_{\text{samples}}}$ \ {\scriptsize (compute energies)}\;
    $\{\tilde{p}_i\}_{i=1}^{N_{\text{samples}}} = \{\frac{e^{-E_i}}{\sum_{j=1}^{N_{\text{samples}}} e^{-E_j}} \} \ \ $ {\scriptsize (use softmax to compute probability)}\;
        $\{\tilde{\bfy_i}\}_{i=1}^{N_{\text{samples}}} \gets  \ \sim \text{GMM}(N_{\text{samples}},\{\tilde{p}_i\}_{i=1}^{N_{\text{samples}}},\{\tilde{a_i}\}_{i=1}^{N_{\text{samples}}})$  {\scriptsize (resample with replacement)}\;
        $\{\tilde{a_i}\}_{i=1}^{N_{\text{samples}}} \gets \{\tilde{a_i}\}_{i=1}^{N_{\text{samples}}} + \ \sim \mathcal{N}(0,\sigma)$ {\scriptsize(add random noise)}\;
        $\sigma \gets K\sigma $ {\scriptsize (shrink sampling scale)} \;

  }
 $\hat{\bfy} = \arg\max(\{\tilde{p}_i\}, \{\tilde{a}_i\}$)
 \caption{Derivative-Free Optimizer}
\end{algorithm}


\subsubsection{Langevin MCMC}
 We introduce a gradient-based MCMC training which uses gradient information for effective sampling and initializes chains from random noise for more mixing. Given an action datapoint $a$, let $E_\theta(a)\in\mathcal{R}$ be the energy function. The energy function would be represented by any function approximator but in our experiments conducted it is represented by a neural network with parameters $\theta$. The energy function is used in tandem with the Boltzmann distribution $p_\theta(x) = \frac{\exp(-E_\theta(x))}{Z(\theta)}$, where $Z(\theta) = \int\exp(-E_\theta(x))dx$
denotes the partition function.  We further maintain a replay buffer of past samples and use them to initialize Langevin dynamics to allow mixing between chains.
 Specifically, 
 \[
 {}^k\tilde{a}^j_i =
{}^{k-1}\tilde{a}^j_i -\lambda(\frac{1}{2}\nabla _aE_\theta(o_i,
{}^{k-1}\tilde{a}^j_i
)+\omega^k,\; \omega^k \sim \mathcal{N}(0, \sigma)
\]
where  ${}^k\tilde{a}^j_i$ means the $j-$th negative action sample in $i-$th batch of the $k-$th iteration. We let the above iterative procedure define a distribution $q_\theta$ such that $\tilde{a}^k\sim q_\theta$. \cite{welling2011bayesian} showed that $q_\theta \rightarrow p_\theta$ almost surely as $k\rightarrow\infty$. Thus, samples are generated implicitly by the energy function $E_\theta$ as opposed to being explicitly generated by a neural network.


\section{Experiments}
\begin{figure}[t]
    \begin{minipage}{0.6\linewidth}
		\centering
         \includegraphics[width=90mm]{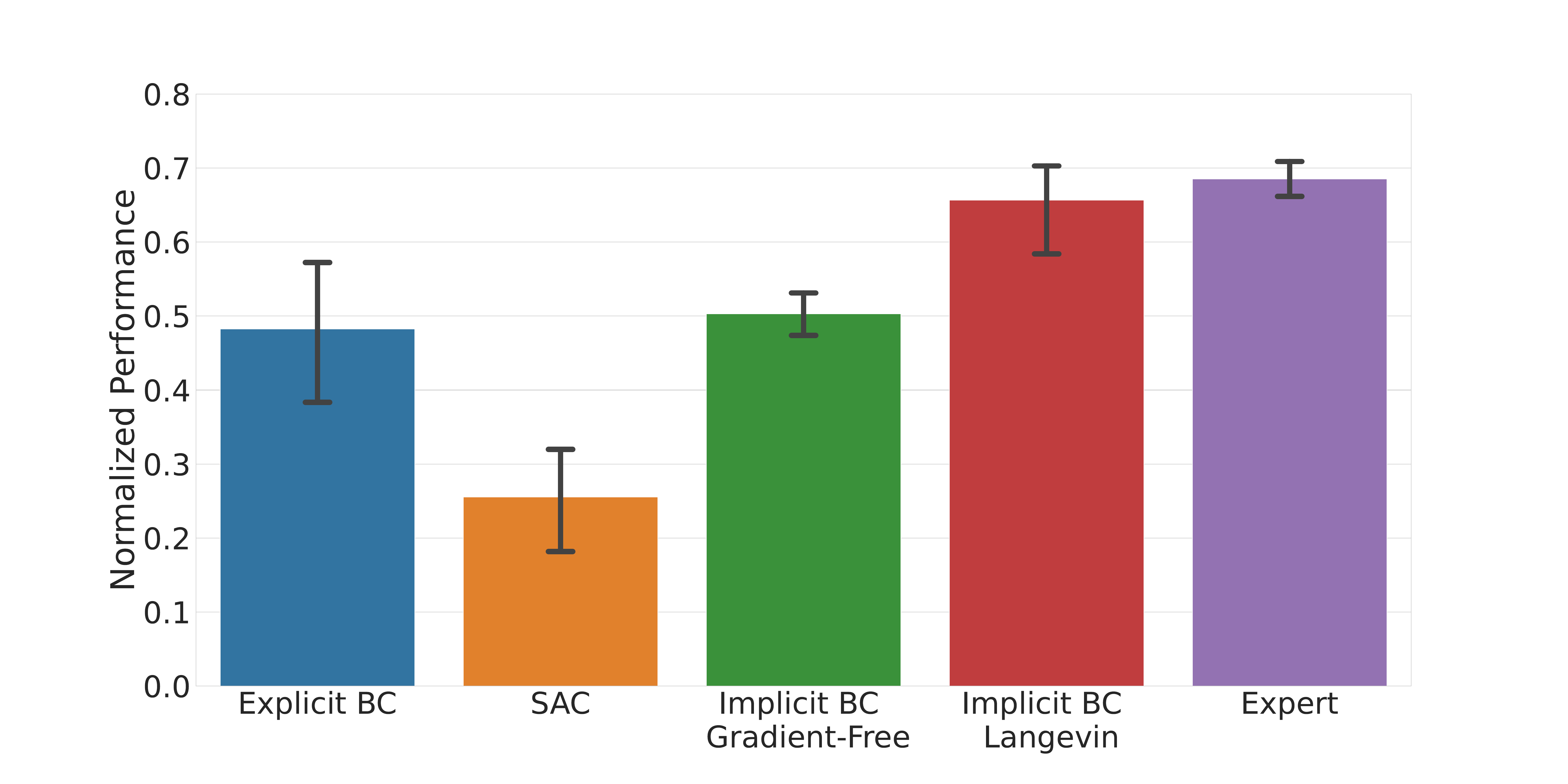}
	\end{minipage}
	\begin{minipage}{0.3\linewidth}
		\centering
		\begin{tabular}{|c|c|}
            \hline
			\textbf{Method} & \textbf{Performance}  \\
            \hline
			Explicit BC     & $0.48 \pm 0.15$  \\
			\hline
			SAC  & $0.25 \pm 0.12$  \\
			\hline
			IBC (grad-free) & $0.50 \pm 0.05$ \\
			\hline
			IBC (Langevin) & $\mathbf{0.65 \pm 0.10}$ \\
			\hline
			Expert & $\mathbf{0.69 \pm 0.05}$ \\
			\hline
		\end{tabular}
		\label{table:student}
	\end{minipage}\hfill
     \caption{Normalized performance on 10 held out configurations.}
     \label{fig:mainresult}
\end{figure}
\subsection{Experiments setup} 
\textbf{Task.} We conduct our simulation experiments in PlasticineLab \cite{huang2021plasticinelab}, which provides a differentiable simulator that can simulate elastoplastic materials such as dough. Given a dough in a spherical shape, our task is to use a cylindrical roller to flatten the dough into a circular shape. We vary the initial and target dough locations, initial roller location, as well as the volume of the dough. An example of a trajectory in our simulation environment is shown at the top of Figure~\ref{fig:traj}. 

\textbf{Evaluation metric.} We use the normalized final Earth Mover's Distance (EMD) as our performance metric in simulation, defined as: 
$$
    \frac{D_{EMD}(P^d_{0}, P^d_{g}) - D_{EMD}(P^d_{T}, P^d_{g})} { D_{EMD}(P^d_{0}, P^d_{g})}
$$ 
where $P^d_{0}$, $P^d_{T}$, $P^d_{g}$ are the ground-truth dough point clouds at initialization, final timestep, and the target, respectively. 

\textbf{Baselines.} We consider several baselines in simulation. First, we compare our method with a  model-free RL baseline trained with point clouds as input: Soft Actor Critic (SAC) \cite{haarnoja2017soft}. Both the actor and the critic in SAC use the same inputs and the same encoder as our method. Specifically, we input the partial point cloud with a PointNet++ \cite{qi2017pointnetplusplus} encoder and then perform standard actor-critic reinforcement learning with the latent feature. We train the SAC agent for 1 million timesteps and average the performance over $4$ random seeds. Second, we compare with a explicit BC policy that directly outputs the control of the roller. Last, we compare the two different version of the implicit BC method. One trained on random negative samples and the other trained with gradient-based, Langevin MCMC as in \cite{florence2021implicit} and \cite{du2019}.

\subsection{Implementation details}
We use our trajectory optimizer to generate $150$ demonstration trajectories uniformly sampled over $125$ initial and target configurations. The trajectory optimizer runs Adam Optimizer~\cite{adam} for $1000$ steps with a learning rate of $0.005$. We then add Gaussian noise $\epsilon \sim \mathcal{N}(0, 0.01)$ to the demonstrations during to prevent overfitting in training. Our implicit model first maps the input point clouds to a compact state representation as a $1024$-dimensional vector and then contact with the action to output an energy value.

\subsection{
Results
}
We evaluate all methods on 10 held-out configurations that have unseen initial and target states from training. First, we see that the trajectory optimizer (``Expert") outperforms the SAC baseline by a wide margin, highlighting the advantage of a differentiable simulator when the state space is high-dimensional. We also see that both versions of the implicit policies trained on the demonstration data outperforms the explicit BC agent, showing again the benefit of learning an energy-based model over the state and action space. Last, utilizing the gradient information in both training and sampling, implicit BC with Langevin Dynamics is the best-performing method for this task.
\begin{figure}[h]
    \centering
    \includegraphics[width=0.6\linewidth]{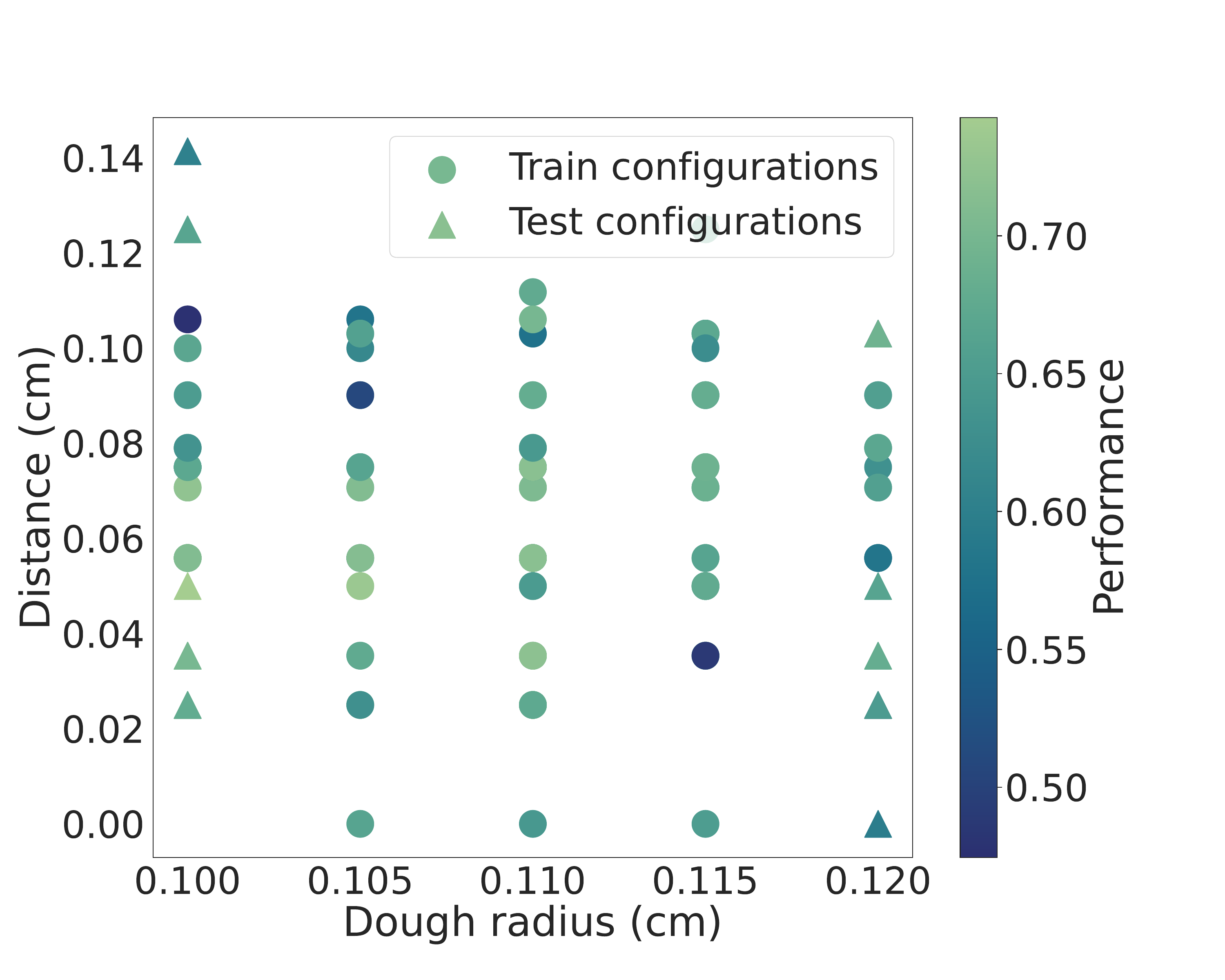}
    \caption{Performance of implicit BC with Langevin Dynamics over all configurations in the demonstration data. The policy is robust to unseen configurations (triangles).
    }
    \label{fig-heatmap}
\end{figure}

To further demonstrate the generalization power of a implicit BC policy, we show the performance of implicit BC with Langevin Dynamics on both training and held-out configurations  in Figure~\ref{fig-heatmap}. Each point in the figure represents the performance of our policy on a specific configuration in the demonstration data. The policy is trained on points represented by cirles and tested on the points represented by triangles. Although the test set contains more extreme values of dough size and target distance, our policy generalizes well to those configurations.

\section{Conclusion}
Compared to explicit regression model, implicit energy model can better approximate discontinuous, multi-valued function. In our dough rolling task, the optimal action is a two-stage process where the rolling pin need to first move down and contact with the dough, and then move back-and-forth horizontally to shape the dough into goal configuration. Our experiments suggest that implicit policy can better approximate this discontinuous trajectory and also better capture the multi-modality of actions, e.g., the rolling pin can move in different directions to flatten a dough.

For training the energy-based model, although random negative samples can provide useful learning signal, this procedure can be inefficient, especially when the dimension of the state space is large. In our case, we find that using a MCMC with Langevin Dynamics can effectively produce better negative samples and result in increase in performance and generalization. 

Last, for inference, a derivative-based sampler is also more effective and yields better optimization results. This is related to the fact that our action space is relatively high dimensional as well. Similar phenomena can also be found in experiments in \cite{florence2021implicit}.



\bibliographystyle{abbrvnat}
\bibliography{ref}
\end{document}